# Assessing Functional Neural Connectivity as an Indicator of Cognitive Performance[*]

Brian S. Helfer[1], James R. Williamson[1], Benjamin A. Miller[1], Joseph Perricone[1], Thomas F. Quatieri[1]

MIT Lincoln Laboratory, 244 Wood Street, Lexington MA USA
`[brian.helfer, jrw, bamiller, joey.perricone, quatieri]@LL.mit.edu`

**Abstract.** Studies in recent years have demonstrated that neural organization and structure impact an individual's ability to perform a given task. Specifically, individuals with greater neural efficiency have been shown to outperform those with less organized functional structure. In this work, we compare the predictive ability of properties of neural connectivity on a working memory task. We provide two novel approaches for characterizing functional network connectivity from electroencephalography (EEG), and compare these features to the average power across frequency bands in EEG channels. Our first novel approach represents functional connectivity structure through the distribution of eigenvalues making up channel coherence matrices in multiple frequency bands. Our second approach creates a connectivity network at each frequency band, and assesses variability in average path lengths of connected components and degree across the network. Failures in digit and sentence recall on single trials are detected using a Gaussian classifier for each feature set, at each frequency band. The classifier results are then fused across frequency bands, with the resulting detection performance summarized using the area under the receiver operating characteristic curve (AUC) statistic. Fused AUC results of 0.63/0.58/0.61 for digit recall failure and 0.58/0.59/0.54 for sentence recall failure are obtained from the connectivity structure, graph variability, and channel power features respectively.

## 1 Introduction

Recent studies have investigated the efficiency and cost of brain functional networks as a way of assessing cognitive aptitude and performance [1][2][3][4][5][6]. Cognitive performance is positively correlated with network communication efficiency [1][6], but this efficiency must be weighed against wiring and energy costs [2][4][7]. An avenue for investigating these tradeoffs and their implications for cognitive per-formance is to compare functional network connectivity properties to performance levels on different cognitive tasks [1][8].

Successful performance in demanding cognitive tasks requires coordinated neural activity in functional brain networks operating at different frequency bands [9][10]. Therefore, to detect cognitive performance failures on a working memory task, we evaluate functional networks that are inferred from EEG at multiple frequency bands. In this paper we assess two novel approaches that characterize the distributional properties of functional networks, and provide a comparison of these approaches with a non-network approach that is based on multichannel EEG power [11]. In contrast to many previous studies, we evaluate properties of brain networks during task execution, thereby indicating the potential for detecting cognitive failures in real time.

## 2    Materials

In this work we use a previously collected database described in [12]. This database provides an IRB-approved auditory working memory protocol where, in each trial, subjects listen to a series of digits followed by a sentence. After a one-second retention interval the subjects are required to repeat the sentence followed by the digits. Recall accuracy was identified based on correct repetition of the entire sentence (sentence recall) and series of digits (digit recall). Each subject was presented with 108 trials at each of three different load levels (defined by the number of digits) and these levels were determined on a per subject basis during an initial calibration procedure. Among the 16 subjects who had EEG data collected, there were 14 subjects who all had a common absolute load level of four digits. Ability level varied across subjects, leading to four digits being a difficult condition for some, and an easy condition for others. The uniform digit setting analyzed in this paper avoids a potential confound due to changes in the task. Our data set consists of 1512 trials (14 subjects with 108 trials per subject), containing 388 digit recall failures and 243 sentence recall failures.

During the working memory protocol, speech, video data, physiology, and electroencephalography (EEG), were all collected. This paper presents an analysis of the predictive performance of EEG in identifying sentence and digit recall failure on a per trial basis. EEG was used from the combined sentence listening/retention period, which varied in duration from 3.8 to 7.7 seconds (mean = 5.3 seconds). The EEG data were preprocessed with a highpass filter at 0.1 Hz, a notch filter at 60 Hz, and had blinks removed using independent component analysis.

## 3    Methods

### 3.1    Coherence as a Measure of Functional Connectivity

In order to obtain sensitive multivariate measures of frequency-band dependent communication between brain regions, we utilize features based on the full set of channel pairwise coherence at four frequency bands implicated in working memory (theta, alpha, beta, gamma) [9][10]. The coherence between channels indicates the amount of cross-channel power in a frequency band relative to the amount of within-channel power. This provides a measure of how closely related the signals are in a frequency band, and therefore a likelihood of brain communication at that band.

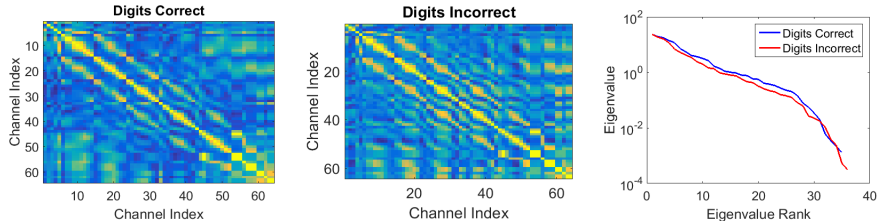

**Fig. 1.** Coherence matrices for digit correct (left) and incorrect (middle) recall trials. Rank-ordered eigenvalues from these trials (right).

Specifically, channel-pairwise coherences are computed to yield a 64×64 coherence matrix for each frequency band in each trial. Figure 1 shows, for example, theta coherence matrices from two nearby trials in the middle of a subject's 2-hour session. The matrices correspond to correct digit recall (left) and incorrect recall (middle). There is a striking difference in the appearance of the two coherence matrices, with the incorrect recall matrix containing a greater number of strong pairwise coherences.

### 3.2 Connectivity Structure

To obtain features that are sensitive to the overall structure of functional connectivity among EEG channels while being invariant to the particular channel identities, we propose novel connectivity structure features. Connectivity structure at a frequency band refers to the distribution of eigenvalues in a coherence matrix from that band, which encodes the overall shape of the multivariate coherence distribution. Figure 1 (right) plots the connectivity structure features, which are the rank-ordered eigenvalues from the two coherence matrices, illustrating that the digit-correct case (blue) produces greater power in the mid-level matrix eigenvalues than the digit-incorrect case (red). This indicates a more isotropic pattern of coherences during correct performance, signifying a greater level of complexity in cortical communication.

To identify how the differences in Figure 1 generalize across trials and frequency bands, we first normalize (z-score) the eigenvalues at each rank across all the trials in the data set, and then plot the average of the normalized eigenvalues for the digit-correct (blue) and digit-incorrect (red) cases in Figure 2. We see that a similarly large difference is found between the correct and incorrect cases in all frequency bands. In addition, we observe a rightward shift in the eigenvalue differences as we move to higher frequency bands. The functional implication of the shifting pattern of these structural connectivity features over frequency bands warrants further investigation.

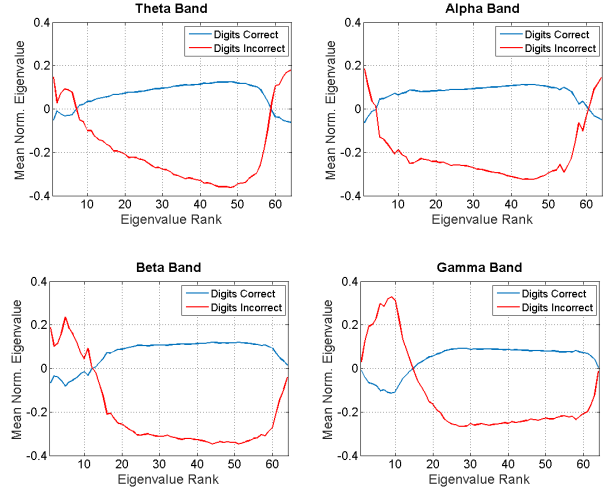

**Fig. 2.** Differences in normalized connectivity structure measures for digit correct (blue) and digit incorrect (red) recall in four frequency bands.

### 3.3 Graph Measures of Functional Connectivity

Much recent work has used graph-based measures of functional brain connectivity. In this section we consider measures that characterize the variability of connectivity patterns in graphs that are constructed based on a range of wiring costs. Each graph consists of a set of nodes (EEG channels) with edges representing connections (high coherence) between the nodes. A threshold on pairwise coherence is used to determine the existence of an edge. For each frequency band, we create four graphs that vary in wiring cost by choosing coherence thresholds that produce the average number of edges per node (node degree) of 6, 6.5, 7, and 7.5. Then, we obtain statistical measures of within-graph variability as described below.

Within-graph variability is measured using two graph features that are implicated in network efficiency and "small world" structure: 1) *average path length (APL) of connected components* and 2) *degree*. A node's average path length is the expected value of the minimum path length to a node selected uniformly at random from the set of nodes for which such a path exists. Average path length was computed over connected components, which were defined as a set of nodes that could be reached by at least one path in the network. A node's degree is simply its number of edges. Of the statistical measures that were considered (max, min, mean, and standard deviation across the 64 nodes), standard deviation of the graph features provided the greatest discriminative ability for detecting recall failure. So, for each frequency band we produce an 8-dimensional vector comprising the standard deviation of the graph features in networks constructed at four different cost levels. To depict these variability measures as a function of frequency band, we averaged the measures across cost levels and then plotted the z-scored measures for correct (blue) and incorrect (red) digit recall in Figure 3.

For both measures a crossover in the patterns occurs as we move from low to high frequency, with path length variability for digits-incorrect trials being lower for theta but higher in the other bands. Degree variability for digits-incorrect is higher in theta and alpha, but lower in beta and gamma. The greater path length variability combined with lower degree variability for digits-incorrect trials implies that failure is associated with a less complex network topology in the higher frequency bands.

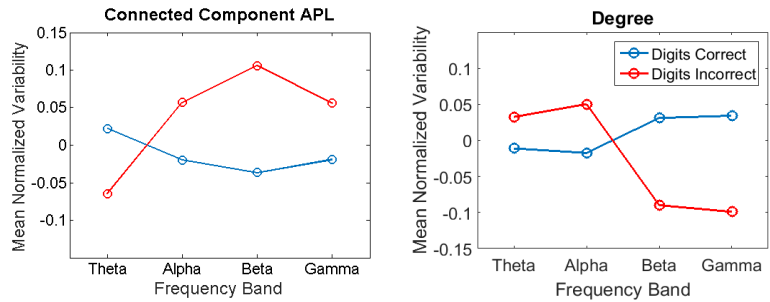

**Fig. 3.** Differences in normalized graph variability measures as a function of frequency band for digit correct (blue) and digit incorrect (red) recall.

## 4  Machine Learning Approach

### 4.1  Dimensionality Reduction via Principal Component Analysis

Connectivity properties and power features were extracted to detect recall failure for digits and sentences on a trial-by-trial basis. Statistical models were trained using these features in cross-validation on 13 subjects and applied to a held-out test subject. However, before training the models we performed dimensionality reduction on each feature set. There is a separate feature set for each feature type at each frequency band. Thus, there are 64 connectivity structure, eight graph variability, and 64 log-power features in each of four frequency bands. We employ principal component analysis (PCA) to reduce dimensionality, with PCA coefficients derived from z-scored training set features and then applied to the test set. For all feature sets we use a number of components that explains 90% of the variance, resulting in (across the four frequency bands): 6, 5, 5, 4 connectivity structure features; 3, 3, 2, 2 graph variability features; and 10, 4, 10, 10 log-power features.

### 4.2  Detecting Digit and Sentence Recall Failure

Multivariate Gaussian distributions are used to model the outcome for each feature set. Gaussian classifier output in a trial is the log-likelihood ratio of recall failure versus success. A common covariance matrix, based on the total data covariance across both classes, is used to obtain better regularization on the graph variability features. Results for each feature set are summarized using the area under the receiver operating characteristic (ROC) curves, or AUC. Fusion across frequency bands and feature types is obtained based on summation of log-likelihood ratios. As log-power performs poorly in theta and alpha, it is fused only over the beta and gamma bands.

## 5  Experimental Results

Table 1 displays the effectiveness of the feature sets in detecting digit recall failure and sentence recall failure. For digit recall, all three feature types perform reasonably well in beta and gamma, but only connectivity structure performs well in theta. For sentence recall, log-power fails at theta and performs only moderately well at the other frequency bands, whereas connectivity structure fails only in gamma and graph variability fails only in theta. Fusion across all three feature types was also done, resulting in AUC = 0.67 for detecting digit recall failure and AUC = 0.60 for detecting sentence recall failure.

**Table 1.** Area under ROC curves for connectivity structure, graph variability, and EEG log-power features at four EEG frequency bands, and for results combined across frequency bands (* p-value less than 0.05, ** p-value less than 0.01; Wilcoxon rank sum test).

| Freq Band | Digit Failure AUC | | | Sentence Failure AUC | | |
|---|---|---|---|---|---|---|
| | Connect. Structure | Graph Var. | EEG Power | Connect. Structure | Graph Var. | EEG Power |
| Theta | 0.62** | 0.49 | 0.51 | 0.57** | 0.49 | 0.48 |
| Alpha | 0.57** | 0.54** | 0.45 | 0.57** | 0.54 | 0.53 |
| Beta | 0.60** | 0.57** | 0.60** | 0.53 | 0.57** | 0.54 |
| Gamma | 0.58** | 0.54* | 0.59** | 0.51 | 0.59** | 0.54* |
| Comb. | 0.63** | 0.58** | 0.61** | 0.58** | 0.59** | 0.54 |

## 6  Conclusion

In this paper we examined the use of novel measures of neural connectivity and power derived from EEG for the real-time detection of auditory working memory failure, and showed the discriminative ability of all three feature types. Previous experimental work has implicated the need for long-range neural communication in specific frequency bands to perform many specific working memory tasks [9][10]. Of particular interest here is the finding of theta-gamma involvement in auditory working memory [9]. All three feature types that we examined contributed to digit recall detection in the gamma band, but only the connectivity structure features supported detection in the theta band. Recall failure appears to be associated with lower complexity in connectivity structure and graph variability. In future work we will expand our current efforts by considering connectivity based on cross-frequency couplings among brain regions [9] and by assessing more complex and informative measures of graph connectivity.